# Meta-aprendizado para otimização de parâmetros de redes neurais.

Tarcísio D. P. Lucas[1], Teresa B. Ludermir[1], Ricardo B. C. Prudêncio[1], Carlos Soares[2]

[1]Centro de Informática – Universidade Federal de Pernambuco (UFPE)
Av. Jornalista Anibal Fernandes – Cidade Universitária – 50740-560 – Recife – PE – Brasil

[2]Faculty of Economic
University of Porto (LIACC) – Porto – Portugal

{tdpl,rbcp,tbl}@cin.ufpe.br, csoares@fep.up.pt

**Abstract.** *The optimization of Artificial Neural Networks (ANNs) is an important task to the success of using these models in real-world applications. The solutions adopted to this task are expensive in general, involving trial-and-error procedures or expert knowledge which are not always available. In this work, we investigated the use of meta-learning to the optimization of ANNs. Meta-learning is a research field aiming to automatically acquiring knowledge which relates features of the learning problems to the performance of the learning algorithms. The meta-learning techniques were originally proposed and evaluated to the algorithm selection problem and after to the optimization of parameters for Support Vector Machines. However, meta-learning can be adopted as a more general strategy to optimize ANN parameters, which motivates new efforts in this research direction. In the current work, we performed a case study using meta-learning to choose the number of hidden nodes for MLP networks, which is an important parameter to be defined aiming a good networks' performance. In our work, we generated a base of meta-examples associated to 93 regression problems. Each meta-example was generated from a regression problem and stored: 16 features describing the problem (e.g., number of attributes and correlation among the problem's attributes) and the best number of nodes for this problem, empirically chosen from a range of possible values. This set of meta-examples was given as input to a meta-learner which was able to predict the best number of nodes for new problems based on their features. The experiments performed in this case study revealed satisfactory results.*

**Resumo.** *A otimização de Redes Neurais Artificiais (RNAs) é uma tarefa importante para o sucesso do uso desses modelos em aplicações reais. As soluções para essa tarefa em geral são custosas, envolvendo procedimentos de tentativa e erro ou conhecimento especializado nem sempre disponível. Nesse trabalho, investigamos o uso de meta-aprendizado para otimização de parâmetros de RNAs. Meta-aprendizado é uma área de pesquisa que tem como objetivo adquirir conhecimento de forma automática relacionando características dos problemas de aprendizado com o desempenho dos algoritmos candidatos. As técnicas de meta-aprendizado foram originalmente propostas e avaliadas para o problema de seleção de algoritmos e posteriormente para otimização de parâmetros de Support Vector Machines. No entanto, meta-aprendizado pode ser adotada* *como uma estratégia mais geral para otimização de parâmetros de RNAs, o que motiva novos esforços nessa direção. No presente trabalho, realizamos um estudo de caso usando meta-aprendizado para a escolha do número de neurônios ocultos de redes MLP, que é um parâmetro importante a ser definido visando um bom desempenho das redes. No nosso trabalho, construímos uma base de meta-exemplos associada a 93 problemas de regressão. Cada meta-exemplo foi gerado a partir de um problema de regressão e armazenou: 16 características descrevendo o problema (e.g. número de atributos e correlações entre atributos do problema) e o melhor número de neurônios para o problema, escolhido empiricamente de um conjunto de valores possíveis. Esse conjunto de meta-exemplos foi fornecido como entrada para um meta-aprendiz que foi capaz de predizer o melhor número de neurônios para novos problemas a partir de suas características. Os experimentos realizados nesse estudo de caso apresentaram resultados satisfatórios.*

## 1. Introdução

Redes Neurais Artificiais (RNAs) vêm sendo utilizadas na literatura para resolver problemas nas mais diferentes áreas do conhecimento [Braga et al. 2007]. Apesar do potencial desses modelos, seu desempenho é dependente da escolha dos seus parâmetros de treinamento e arquitetura.

A definição dos parâmetros depende do problema (ou tarefa de aprendizagem) sendo resolvida. Considere por exemplo a definição do número de neurônios da camada escondida de redes *Multilayer Perceptron* (MLP). Se esse número for exageradamente pequeno, pode ser insuficiente para a rede neural aprender os padrões existentes nos dados (i.e. *underfitting*). Já se o número de neurônios for exageradamente grande, a rede pode decorar o conjunto de treinamento (i.e. *overfitting*), perdendo sua capacidade de generalização [Braga et al. 2007].

A otimização dos parâmetros de RNAs é um problema complexo e várias soluções já foram propostas para resolvê-lo. O método manual de tentativa e erro é a mais simples, no entanto, esse método pode exigir muito do usuário e pode não levar a bons resultados. Abordagens mais sofisticadas e sistemáticas, como o uso de algoritmos de busca meta-heurística, têm sido bastante exploradas [Zanchettin et al. 2011] [El-Henawy et al. 2010] [Bin et al. 2010] [Li et al. 2010] [Carvalho and Ludermir, 2006]. Nesses casos, os melhores parâmetros são definidos por um processo de otimização, em que o desempenho da rede neural corresponde à função objetivo da otimização, e o conjunto de configurações possíveis de parâmetros corresponde ao espaço de busca. Uma limitação dessa abordagem é que o conhecimento adquirido ao lidar com um problema não é usado para ajudar a resolver novos problemas. Isso faz com que a busca comece sempre de soluções definidas aleatoriamente do espaço de busca para um dado novo problema, levando a uma convergência mais lenta.

Meta-aprendizado é uma área que tenta extrapolar o conhecimento adquirido em problemas passados para resolver problemas novos. Meta-aprendizado pode ser definido como o processo de aquisição automática de conhecimento que relaciona o desempenho dos algoritmos de aprendizado com as características dos problemas [Giraud-Carrier et al. 2004]. O conhecimento em meta-aprendizado é adquirido a partir de um conjunto de meta-exemplos. Em geral, cada meta-exemplo é associado a um problema de aprendizado e é formado por: (1) meta-atributos que descrevem o problema (e.g., número de atributos, número de exemplos, entropia de classe,...) e (2) meta-rótulo (rótulo que indica o algoritmo que obteve melhor desempenho no problema) [Prudêncio and Ludermir 2009]. Diversos trabalhos em meta-aprendizado têm sido desenvolvidos para seleção de algoritmos com sucesso em diferentes domínios de aplicação [Smith-Miles 2008].

As técnicas de meta-aprendizado, originalmente propostas e validadas para seleção de algoritmos, têm sido exploradas para otimização de parâmetros, mais especificamente para otimização de parâmetros de *Support Vector Machines* (SVM) [Alia and Smith-Miles 2006] [Soares et al. 2004] [Prudencio et. al 2004]. O bom desempenho do meta- aprendizado nesses trabalhos motiva o uso dessa abordagem para outros modelos de RNAs. Neste trabalho, investigamos o uso de meta-aprendizado para otimização de parâmetros de RNAs. Dessa forma, desenvolvemos um estudo de caso que aplica meta- aprendizado para escolha do número de neurônios escondidos para redes MLP com uma camada escondida. Para tal, construímos uma base de meta-exemplos a partir da avaliação de redes MLP em 93 problemas de regressão, variando sistematicamente o número de neurônios da camada escondida. Cada meta-exemplo é associado a um problema de regressão e armazena (1) 16 características descrevendo os dados dos problemas; (2) o melhor número de neurônios para a rede MLP no problema, definido de forma empírica. Um algoritmo de regressão usado como meta-aprendiz recebeu esse conjunto de meta- exemplos como entrada, sendo usado para predizer o melhor número de neurônios para um novo problema a partir de suas caracter´ısticas. No nosso trabalho, avaliamos quatro algoritmos de regressão diferentes: regressão linear, k-NN, M5 e SVM. O estudo de caso validou o meta-aprendizado como alternativa na otimização do número de neurônios da camada escondida. Verificamos nos experimentos que as características dos problemas (meta-atributos) podem ser utilizadas como uma forma de se indicar o número de neurônios da camada escondida de redes MLP.

O restante desse artigo é organizado da seguinte forma. Na Seção 2, é feita uma revisão bibliográfica sobre o problema de otimizar RNAs. A Seção 3 mostra os conceitos básicos de meta-aprendizado. Em seguida, na Seção 4, o trabalho desenvolvido é detalhado. Por fim, na Seção 5, encontram-se as conclusões e trabalhos futuros.

## 2. Otimização de redes neurais

Redes Neurais Artificiais (RNAs) vêm sendo utilizadas na literatura para resolver problemas nas mais diferentes áreas de conhecimento [Mar et al. 2011] [Gomperts et al. 2011] [Patra and Chua 2010] [Achkar and Nasr 2010] [Harun et al. 2010] [Ferreira and Ludermir, 2009]. O sucesso desse tipo algoritmo em aplicações motivou o desenvolvimento crescente de novos modelos com diferentes níveis de complexidade. No entanto, a qualidade de um modelo neural é fortemente dependente da definição adequada dos valores de seus parâmetros (e.g., tipo de rede, número de camadas, número de neurônios, taxas de aprendizagem, dentre outros). Para um dado novo problema, existe uma parametrização ótima e específica a ser aplicada no modelo neural. Definir os valores de tais parâmetros não é uma tarefa fácil, pois o número de configurações possíveis costuma ser muito alto. Além disso, existe um custo computacional para avaliar o quão boa é certa parametrização, que varia de acordo com o problema, modelo neural e algoritmo de treinamento escolhido. Obviamente, uma busca exaustiva no espaço de parâmetros não é viável na prática.

Existem várias abordagens para se otimizar os parâmetros de uma RNA. Em muitos casos, a otimização dos parâmetros é feita por tentativa e erro, de forma não sistemática pelo usuário das redes. Essa abordagem pode não levar a bons resultados se o usuário tiver pouca experiência para guiar o processo de busca. Nesse contexto, diferentes autores têm tratado a otimização de RNAs como uma tarefa de otimização, adotando algoritmos de busca meta-heurística para explorar sistematicamente o espaço de configurações possíveis de parâmetros. Diferentes algoritmos já foram usados nesse contexto como Simulated Annealing, Tabu Search, Algoritmos Genéticos e Otimização por Enxame de Partículas [Zanchettin et al. 2011] [El-Henawy et al. 2010] [Bin et al. 2010] [Li et al. 2010] [Minku and Ludermir, 2008].

A Figura 1 ilustra o uso de um algoritmo de busca para otimizar parâmetros para uma rede MLP. Os algoritmos de busca partem de uma ou mais soluções normalmente geradas aleatoriamente. No caso da Figura 1, são geradas várias soluções, cada uma armazenando dois parâmetros: número de neurônios ocultos (*ne*) e taxa de aprendizagem (*ta*). Cada solução é ainda associada a um valor de *fitness* que nesse caso corresponde ao desempenho empírico (e.g., taxa de acerto estimada por validação cruzada) da rede neural quando os parâmetros são adotados. Operadores de busca são adotados iterativamente sobre as soluções atuais para encontrar novas soluções que otimizem a função objetivo desejada.

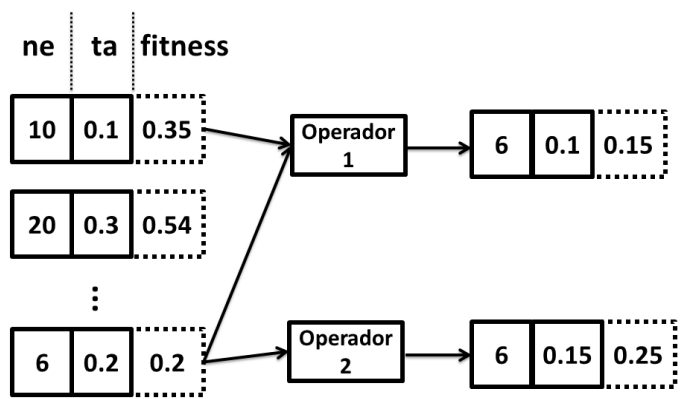

**Figura 1. Modelagem de um algoritmo de busca para selecionar parametrização de uma rede MLP com uma camada escondida, onde *ne*, *ta* e *fitness* representam respectivamente número de neurônios escondidos, taxa de aprendizagem e o retorno da função objetivo para tal.**

Embora tragam bons resultados, tais abordagens ainda são custosas uma vez que, para o cálculo da função objetivo, é necessário avaliar empiricamente todas as configurações de parâmetros geradas pelo algoritmo ao longo da busca. Além disso, os algoritmos de busca têm por sua vez uma quantidade expressiva de parâmetros a definir (e.g., taxas de *crossover* e mutação, tamanho da população e número de gerações), que têm impacto alto na qualidade do processo de busca. Finalmente, citamos que o conhecimento adquirido ao lidar com um problema não é extrapolado para ajudar a resolver outros problemas. Com isso, para cada novo problema, o processo de busca começa sempre de soluções aleatórias e o algoritmo pode demorar a convergir.

Uma alternativa para a otimização de RNAs é o uso de conhecimento sobre os modelos. De fato, existem uma quantidade grande de regras e heurísticas que podem ser usadas para guiar a escolha de alguns parâmetros de modelos específicos [Sarle 1997]. Por exemplo, em [Blum 1992], é sugerido que o número de neurônios ocultos de uma rede MLP seja menor que o número de entradas (i.e., número de atributos do problema) e maior que o número de saídas da rede. Embora essas regras possam ser úteis em alguns contextos, elas são limitadas e pouco extensíveis. De fato, em geral é difícil levar em consideração características mais complexas dos dados como correlações entre atributos e ruído nos dados. Além disso, é difícil elicitar conhecimento para modelos de RNAs mais complexos uma vez que especialistas nem sempre estão disponíveis. Nesse contexto, meta-aprendizado surge como uma alternativa para adquirir automaticamente conhecimento que leva em consideração características dos dados para recomendar os melhores algoritmos e parâmetros. Essa abordagem é menos custosa computacionalmente quando comparada com as técnicas de busca e ao mesmo tempo fornece uma estratégia sistemática para aquisição de conhecimento que pode ser adotada para otimização de parâmetros em diferentes modelos de RNAs.

## 3. Meta-aprendizado

Meta-aprendizado pode ser definido como o processo de aquisição automática de conhecimento, que relaciona o desempenho dos algoritmos de aprendizagem com as características dos problemas [Giraud-Carrier et al. 2004]. O conhecimento em meta-aprendizado é adquirido a partir de um conjunto de meta-exemplos. Cada meta-exemplo, de uma forma geral, é formado por: (1) meta-atributos que definem o problema (número de atributos, número de exemplos, entropia do rótulo, etc.) e (2) meta-rótulo (algoritmo que obteve melhor resultado no problema) [Prudêncio and Ludermir 2009].

Para aplicar meta-aprendizado é necessário construir uma base de meta-exemplos (ou meta-base). Cada meta-exemplo é composto por um conjunto de meta-atributos e um meta-rótulo (variável alvo). Os meta-atributos são, em geral, estatísticas que descrevem o conjunto de dados, tais como número de exemplos, número de atributos, correlação entre atributos, entropia média dos atributos, entre outros [Brazdil et al. 2003] [Kalousis et al. 2004]. Já o meta-rótulo, de uma forma geral, é o melhor algoritmo aplicado ao problema, dentre um conjunto de algoritmos candidatos [Prudêncio and Ludermir 2009]. O atributo alvo é determinado normalmente de forma experimental (por exemplo, validação cruzada).

A maioria dos trabalhos de meta-aprendizado foi proposta para seleção de algoritmos. No entanto, desde 2004, o meta-aprendizado também tem sido aplicado para otimização de parâmetro de algoritmos. Os trabalhos feitos utilizando meta-aprendizado para otimização de parâmetros foram aplicados a máquina de vetores de suporte (do inglês Support Vector Machine - SVM). Num desses trabalhos, tentou-se encontrar a melhor função de *kernel* de um SVM para problemas de classificação [Alia and Smith-Miles 2006]. Nesse caso, a tarefa de meta-aprendizado é uma classificação em que o meta-rótulo é a melhor função *kernel* dentre 5 candidatas, para um determinado problema de classificação. Nesse trabalho foram construídos 112 meta-exemplos, a partir de 112 bases de dados de classificação.

Em outro estudo, procurou-se encontrar a relação entre a largura do *kernel* Gaussiano e os meta-atributos para problemas de regressão [Soares et al. 2004]. Embora o parâmetro de largura do *kernel* Gaussiano seja contínuo, o problema foi tratado como sendo de classificação, tendo como variável objetivo 11 diferentes valores discretos. Nesse trabalho, foram construídos 42 meta-exemplos, a partir de 42 bases de dados de regressão.

Os resultados promissores obtidos na aplicação de meta-aprendizado para otimização de parâmetros de SVMs [Soares et al. 2004] [Alia and Smith-Miles 2006] motivaram a expansão desse tipo de abordagem para otimização de parâmetros de redes neurais. Este trabalho é um estudo de caso que lança a ideia de aplicar meta-aprendizado na otimização dos parâmetros de RNAs. A rede escolhida para tal foi a MLP com uma camada escondida.

## 4. Meta-aprendizado para otimização de redes neurais artificiais

Este estudo de caso investiga a utilização de meta-aprendizado na otimização de parâmetros de redes neurais artificiais (RNAs). A construção da meta-base tem, normalmente, um alto custo computacional. Após a geração da base, ainda é necessário treinar o meta-aprendiz com os meta-exemplos gerados. No entanto, o custo computacional de utilização do modelo desenvolvido é muito baixo. Neste trabalho, a investigação se dará em cima da otimização de um dos parâmetros de uma rede MLP com uma camada escondida. O algoritmo de treinamento utilizado para treinar a rede foi o Extreme Learning Machine (ELM).

O ELM é um algoritmo de treinamento para RNAs MLP com uma camada escondida. Sua principal vantagem, com relação a outros algoritmos de treinamentos clássicos, como o *Backpropagation*, é a velocidade de treinamento, que é extremamente veloz. Outra vantagem é que o ELM possui apenas dois parâmetros: número de neurônios escondidos e função de ativação, que deve ser infinitamente diferenciável. Informações detalhadas sobre o funcionamento do ELM podem ser encontradas em [Huang et al. 2004] e [Huang et al. 2006]. Tais vantagens motivaram o uso do ELM para o estudo de caso. A rapidez do treinamento do ELM torna a construção dos meta-exemplos uma tarefa menos custosa do que se fosse utilizado um algoritmo mais lento, como, por exemplo, o *Backpropagation*.

O parâmetro escolhido para o estudo de caso foi o número de neurônios escondidos da rede MLP. Dessa forma, foi fixada uma função de ativação e testou-se todos os valores de neurônios escondidos no intervalo [1, 300] para 93 diferentes bases de dados de regressão. Foram extraídos também os meta-atributos de cada uma das bases de dados. Com isso, foi possível montar uma meta-base com 93 meta-exemplos. Por fim, alguns algoritmos de regressão foram testados como opções de meta-aprendiz.

### 4.1. Meta-exemplos

A construção dos meta-exemplos passa primeiro pela coleta das bases de dados. Foram selecionadas 93 bases de regressão obtidas através do site do WEKA [1] . Foram selecionadas bases com no mínimo 100 exemplos, e que o atributo alvo (ou rótulo) tivesse pelo menos 10 valores distintos. Tais restrições tiveram como objetivos aumentar a probabilidade de obter problemas que: (1) tivessem conhecimento a ser aprendido (mínimo de 100 exemplos) e que (2) tivessem uma diversidade mínima na variável a ser predita (pelo menos 10 valores distintos)

---
[1]Site do WEKA - http://www.cs.waikato.ac.nz/ml/weka/

Cada uma das 93 bases teve seus dados normalizados. As variáveis objetivo foram normalizadas entre zero e um. Já as variáveis de entrada contínuas foram normalizadas entre -1 e 1. Atributos simbólicos sofreram transformação para representação binária, onde um atributo com *k* diferentes valores passou a ser representados por k-1 bits zeros e um bit 1.

Cada meta-exemplo, no nosso trabalho, foi criado a partir de um problema de regressão e armazenou: (1) 16 meta-atributos descritivos do problema; e (2) o melhor número de neurônios para rede MLP no problema, que será o meta-rótulo. Esses dois aspectos serão definidos a seguir.

### 4.1.1. Meta-atributos

Os meta-atributos coletados tiveram o intuito de representar a complexidade de cada problema. Isso foi feito porque existe uma relação entre a complexidade de um problema e a quantidade de neurônios suficiente para aprender os exemplos que constituem a base. No nosso trabalho foram usados 16 meta-atributos. A escolha dos meta-atributos teve por base trabalhos anteriores de meta-aprendizado aplicado a parametrização do SVMs [Soares et al. 2004] [Alia and Smith-Miles 2006], são eles:

- média de assimetria dos atributos;
- média de curtose dos atributos;
- assimetria média absoluta dos atributos;
- máxima distância média entre cada valor *target* e seus dois vizinhos;
- máxima correlação atributo/*target*;
- número de atributos;
- número de exemplos;
- número de atributos contínuos com *outliers*;
- proporção de atributos contínuos com *outliers*;
- $R^2$ coeficiente de regressão linear múltipla (apenas atributos numéricos são usados);
- $R^2$ coeficiente de regressão linear múltipla (atributos simbólicos são binarizados);
- largura do coeficiente de variação absoluta do *target*: 0 se não escasso, 1 se esparsas e 2 se extremamente escasso;
- largura do coeficiente de variação do *target*: 0 se não escasso, 1 se esparsas e 2 se extremamente escasso;
- existência de *outliers* no atributo alvo: 1 se tiver, 0 caso contrário (calculado para os atributos contínuos);
- valor do *outlier*;
- verificar se o desvio padrão é maior do que a média: 1 se for, 0 caso não seja.

### 4.1.2. Meta-rótulo

Coletados os meta-atributos, o objetivo agora é encontrar o melhor número de neurônios escondidos para as redes MLP em problemas de regressão. No nosso trabalho, utilizamos o algoritmo Extreme Learning Machine (ELM) para o treinamento das redes MLP. O ELM não utiliza conjunto de validação como critério de parada em seu treinamento. Dessa forma, cada base de dados foi dividida em dois conjuntos: treinamento (70% dos dados) e teste (30% dos dados). O conjunto de treinamento foi utilizado para ajustar os pesos da MLP. Já o conjunto de teste foi utilizado para avaliar a precisão do modelo desenvolvido no treinamento. Foram testados todos os valores de neurônios escondidos no intervalo [1, 300]. A função de ativação utilizada foi a *sigmoidal* $g(x) = 1/(1 + exp(-x))$.

Cada experimento foi repetido 10 vezes, com diferentes pesos e bias iniciais. O número de neurônios escondidos considerado ideal para cada problema foi o que obteve o menor erro médio no conjunto de teste nas 10 repetições. O tipo de erro utilizado foi a raiz quadrada média dos erros (do inglês, Root Mean Square Error - RMSE), descrito na Equação 1, onde $n$ é o número de exemplos no conjunto de teste e $P_j$ e $T_j$ são, respectivamente, a saída da rede e a resposta correta (rótulo). Todas as simulações com o ELM foram realizados no MATLAB 7.9, a partir da implementação do ELM do próprio autor[2]. A Tabela 1 ilustra a base de meta-exemplos da meta-base construída. Já a Figura 2 mostra um histograma com a distribuição do meta-rótulo na meta-base. O espalhamento dos valores do meta-rótulo ao longo do histograma indica a complexidade de se aprender o meta-problema de seleção do número ideal de neurônios na camada escondida.

$$RMSE = \sqrt{\frac{1}{n}\sum_{j=1}^{n}(P_j - T_j)^2} \qquad (1)$$

**Tabela 1. Exemplos dos meta-exemplos desenvolvidos para meta-aprendizado do número de neurônios escondidos ideal para um dado problema.**

| Meta-atributos | | | | Meta-rótulo |
|---|---|---|---|---|
| Número exemplos | Número atributos | Número outliers | ... | Neurônios Escondidos |
| 100 | 3 | 0 | ... | 4 |
| 209 | 6 | 5 | ... | 32 |
| 950 | 9 | 0 | ... | 215 |
| ... | ... | ... | ... | ... |
| 13750 | 40 | 0 | ... | 142 |

Desenvolvida a meta-base, a etapa seguinte é submetê-la ao meta-aprendiz para avaliar a precisão da predição do modelo.

### 4.2. Meta-aprendiz

Nesse estudo de caso, a tarefa de meta-aprendizado é prever o número de neurônios da camada escondida ideal para uma rede MLP, baseado nas características do problema ao qual a rede será submetida. O responsável pelo aprendizado dessa tarefa é o meta-aprendiz. Como o meta-rótulo é um inteiro no intervalo [1, 300], optamos por utilizar algoritmos de regressão como meta-aprendiz. O uso de classificadores como meta-aprendiz poderia tornar o problema de meta-aprendizado mais complexo, uma vez que seriam 300 classes a serem aprendidas a partir de apenas 93 exemplos.

---
[2]Site com a implementação do ELM: http://www.ntu.edu.sg/home/egbhuang/

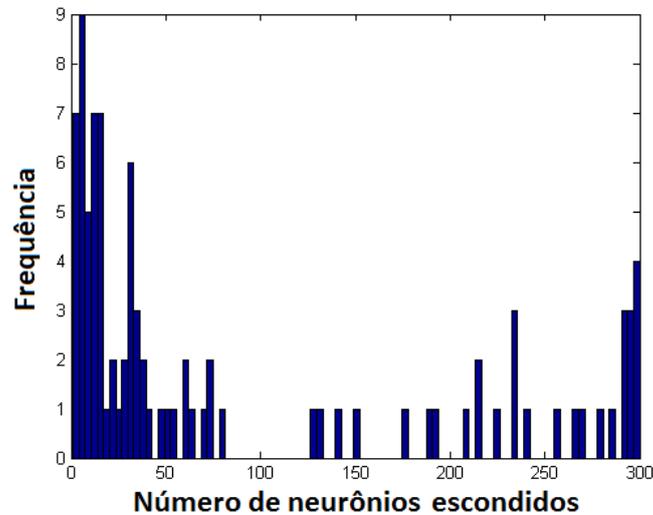

**Figura 2. Histograma que mostra a distribuição das melhores opções de neurônios escondidos para cada um dos 93 problemas de regressão.**

Nesse trabalho, foram testados meta-aprendizes utilizados anteriormente na literatura [Guerra et al. 2008]. Todos os algoritmos estão implementados na ferramenta WEKA, são eles:

- 1-NN: um caso especial do algoritmo de aprendizado baseado em instância (do inglês *Instance-based learning algorithms*) [Aha et al. 1991].
- M5: algoritmo proposto por Quinlan para induzir árvores de regressão [Quinlan 1992].
- Regressão Linear
- SVM: utilizando dois diferentes *kernels*: polinomial e RBF.

Todos os meta-aprendizes foram utilizados na sua parametrização *default* do WEKA, exceto o SVM com *kernel* RBF, onde se testaram outras duas parametrizações baseado nos resultados obtidos com esse meta-aprendiz em [Guerra et al. 2008].

### 4.3. Experimentos e resultados

Nesta seção, é avaliado o desempenho dos meta-aprendizes citados na seção anterior. Para avaliação dos meta-aprendizes, foi adotado um procedimento de *leave one out* com os 93 meta-exemplos disponíveis. Nesse procedimento, cada meta-exemplo é utilizado por vez para testar a precisão do meta-aprendiz, enquanto todos os demais são utilizados em seu treinamento. Dessa forma são feitos 93 treinamentos distintos. A partir desse procedimento, foi observado o erro relativo absoluto (do inglês, relative absolute error - RAE) obtido das respostas dos meta-aprendizes em comparação com os valores reais do melhor número de neurônios em cada problema.

A medida tem valores altos próximos de 100% quando as respostas dos algoritmos têm precisão similar a precisão obtida pela média do atributo-alvo (que foi de 93,8 neurônios escondidos) para os meta-exemplos da base. Nesse caso, valores próximos a

100% indicam que o regressor não está sendo útil e tem desempenho similar a um preditor inocente (predição pela média do atributo-alvo). Além da medida RAE, observamos também a correlação entre as respostas dos meta-aprendizes e as saídas desejadas.

Os resultados obtidos com os meta-aprendizes estão resumidos na Tabela 2. Todos os meta-aprendizes obtiveram RAE abaixo de 100% (a partir de 68,11%). A correlação observada entre as saídas da rede e as respostas corretas variaram entre 0,51 e 0,72. Esses resultados indicam que de fato existem regularidades nos meta-atributos dos problemas que podem ser usados para predizer o número de neurônios ocultos, i.e., que existe conhecimento que pode ser adquirido a partir da base de meta-exemplos. O SVM com função *kernel* obteve a maior precisão e correlação entre os meta-aprendizes testados, com 52,85% de RAE e correlação de 0,72. O SVM já mostrou ser uma boa opção para meta-aprendiz em outros trabalhos de meta-aprendizado, como em [Guerra et al. 2008].

**Tabela 2. Medida (RAE) e coeficiente de correlação para 5 diferentes meta-aprendizes na predição do número ideal de neurônios na camada escondidos em redes MLP.**

| Método | RAE | Correlação |
|---|---|---|
| Regressão Linear | 68,11% | 0,66 |
| SVM (Polinomial) | 60,25% | 0,66 |
| **SVM (RBF)** | **52,84%** | **0,72** |
| 1-NN | 66,26% | 0,51 |
| M5 | 61,73% | 0,62 |

## 5. Conclusão

Este trabalho é um estudo de caso que investiga a aplicação de meta-aprendizado na otimização de redes neurais artificiais (RNAs). A rede neural escolhida para o estudo foi a MLP. O algoritmo de treinamento utilizado no treinamento da MLP foi o *Extreme Learning Machine*. O parâmetro escolhido para otimização foi o número de neurônios escondidos.

Uma meta-base com 93 meta-exemplos foi construída e diferentes meta-aprendizes foram aplicado para aprender o meta-problema proposto. Os erros RAE encontrados pelos meta-aprendizes testados na meta-base foram todos menores que 100%, o que assegura que a otimização do número de neurônios escondidos é um meta-problema.

O melhor erro, como em trabalhos anteriores [Guerra et al. 2008], foi obtido tendo o SVM como meta-aprendiz (52,85%). O alto erro reflete a complexidade do meta-problema proposto. Testes realizados por algoritmos de busca provavelmente obteriam soluções mais precisas. No entanto, o custo de se obter uma resposta a parti do modelo de meta-aprendizado desenvolvido será frequentemente inferior a abordagens com algoritmo de busca. Existe ainda uma alternativa intermediária, que seria utilizar o meta-aprendizado para gerar parametrizações promissoras, e atribuir a um algoritmo de busca a tarefa de refiná-las. Dessa forma, o algoritmo de busca já começaria de uma região promissora, e provavelmente chegaria a uma boa solução num tempo mais satisfatório que a busca iniciada aleatoriamente.

Existem ainda alguns estudos que podem melhorar a precisão obtida com meta-aprendizado. A seleção de meta-atributos foi pouco explorada e pode contribuir nesse sentido. Além disso, outros meta-aprendizes e conjuntos de dados podem ser adicionados a fim de encontrar resultados melhores e mais bem embasados. Outro trabalho futuro seria aplicar meta-aprendizado para otimização de redes neurais que possuem o treinamento mais lento como a MLP treinada com o *Backpropagation*.

## Referências